\pdfoutput=1

\documentclass[11pt]{article}

\usepackage[final]{acl}

\usepackage{times}
\usepackage{latexsym}

\usepackage[T1]{fontenc}

\usepackage[utf8]{inputenc}

\usepackage{microtype}

\usepackage{inconsolata}

\usepackage{graphicx}
\usepackage{amsmath}
\usepackage{multirow}
\usepackage{booktabs}
\usepackage{listings}
\usepackage{enumitem}
\usepackage[table]{xcolor}

\usepackage{booktabs}   
\usepackage{siunitx}    
\usepackage{makecell}   
\title{Orcust: Stepwise‑Feedback Reinforcement Learning for GUI Agent}

\author{
 \textbf{Junyu Lu},
 \textbf{Songxin Zhang},
 \textbf{Zejian Xie},
 \textbf{Zhuoyang Song},
 \textbf{Jiaxing Zhang}
\\
\\
Lionrock AI Lab, China Merchants Research Institute of Advanced Technology
\\
}

\begin{document}
\maketitle
\begin{abstract}
Recent advances in GUI agents have achieved remarkable grounding and action‐prediction performance, yet existing models struggle with unreliable reward signals and limited online trajectory generation. In this paper, we introduce Orcust, a framework that integrates Principle-Constrained Reward Modeling (PCRM) and Online VM-Grounded Trajectory Construction (OVTC) to enhance reasoning reliability and data efficiency in interactive GUI tasks. We leverages environment-verifiable and LLM-derived principle to enforce interpretable reward signals that constrain long chain-of-thought reasoning and rule-based feedback. OVTC spins up instrumented virtual machines to autonomously collect structured GUI interaction trajectories with explicit procedural and structural objectives, enabling the training of a stepwise reward model that robustly captures human preferences and adheres to task-specific constraints. Extensive experiments on standard GUI benchmarks covering perceptual grounding, foundational operations, and end-to-end task execution reveal that Orcust achieves state-of-the-art performance, improving by 22.2\% on ScreenSpot and 23.9\% on ScreenSpot-Pro over the base model (i.e. Qwen2.5-VL-7B). The results demonstrate Orcust's effectiveness in enhancing the reasoning, adaptability and scalability of GUI agents across various environments and task complexities.
\end{abstract}
\section{Introduction}

The rapid proliferation of graphical user interfaces (GUIs) across desktop, web, and mobile platforms has spawned a growing demand for intelligent agents capable of automating complex interaction workflows~\cite{wu2024os-atlas, hong2024cogagent, qin2025ui-tars}. Recent advances in vision-language modeling~\cite{bai2025qwen2.5vl} have empowered agents to perceive interface elements, interpret high-level instructions~\cite{bonatti2024windows-agent, rawles2024androidworld}, and execute atomic actions such as clicks and text entry. However, bridging perception with robust, long-horizon decision-making remains an open challenge, as agents must reason over dynamic layouts and ensure safety and correctness.

\begin{figure}[t]
  \centering
  \includegraphics[width=1.0\linewidth]{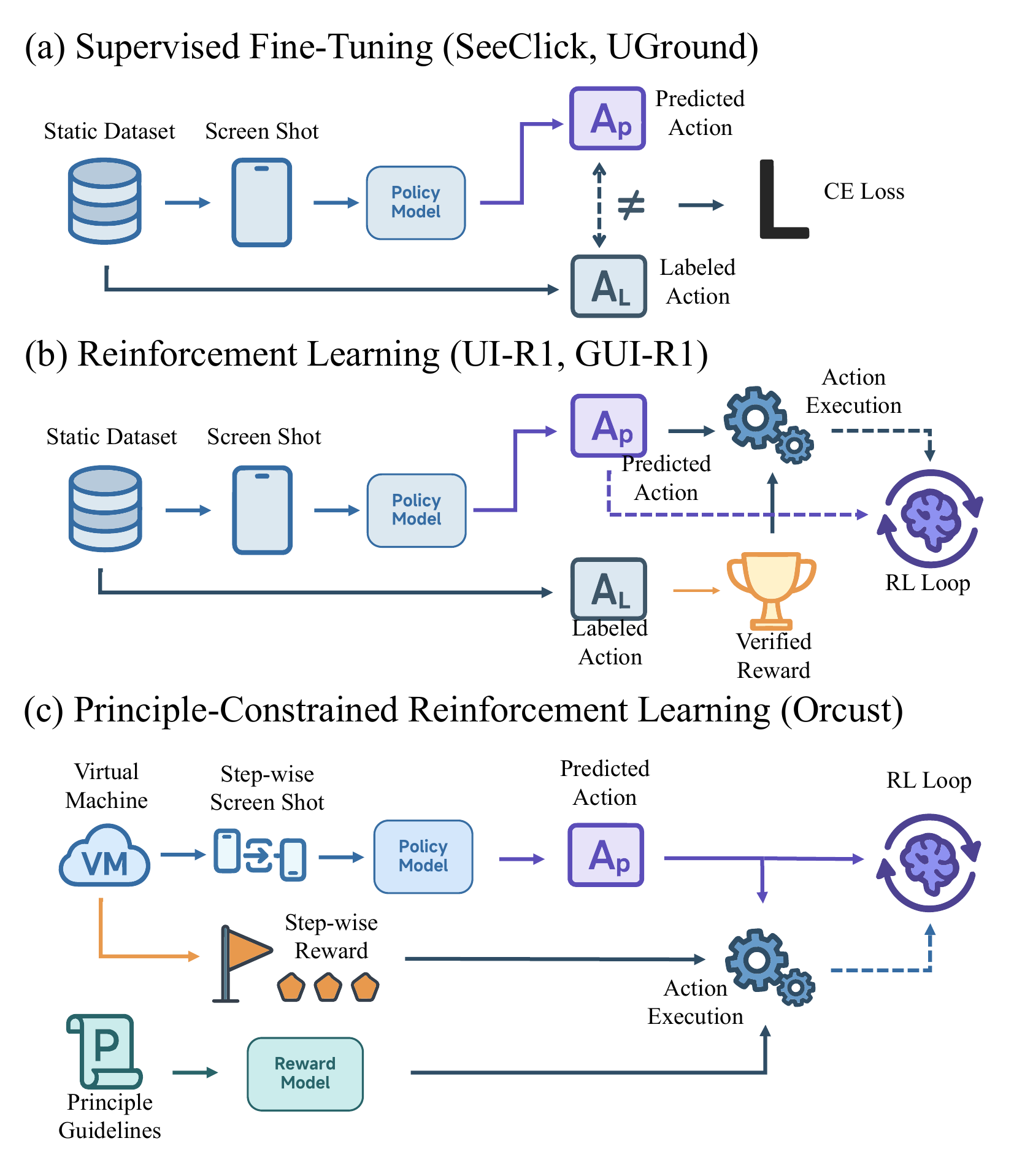}
  \caption{Comparison of GUI agent training approaches. (a) SFT on static, labeled UI interaction data. (b) RFT with trial-and-error in a GUI environment (e.g., rule-based R1 reward shaping). (c) The proposed PCRM that introduces explicit principles and stepwise, verifiable rewards into the RL training loop.}
  \label{fig:compare}
  \vspace{-2mm}
\end{figure}

Prior work on GUI Agents has largely focused on visual grounding and supervised fine-tuning~\cite{gou2024uground, xu2024aguvis, wang2024ponder}. As shown in Figure~\ref{fig:compare} (a), vision-only grounding approaches achieve precise pixel localization of UI components but often lack deep planning capabilities and rely on static, manually annotated datasets. Conversely, supervised action prediction methods deliver strong performance on seen interfaces but struggle to generalize to unseen layouts and require large volumes of labeled data.

Recently, rule-based R1-style reinforcement fine-tuning (RFT) ~\cite{guo2025deepseek-r1, shao2024deepseekmath} and end-to-end models with iterative reflection presented in Figure~\ref{fig:compare} (b) have been introduced to enhance decision making in GUI tasks~\cite{lu2025ui-r1, xia2025gui-r1}. These frameworks integrate structured reward signals and chain-of-thought (CoT) reasoning to improve generalization, but key limitations persist. (i) Rewards are often coarse-grained—success/failure signals delayed until the final screen—hindering credit assignment for long chains of thought. (ii) Existing trajectories are static; they neither evolve with interface updates nor capture emergent failure modes, forcing costly recollection cycles.

In this paper, we propose Orcust, a comprehensive RL framework that delivers robust, sample-efficient learning and stepwise feedback into GUI Agents as Figure~\ref{fig:compare} (c). First, Principle-Constrained Reward Modeling (PCRM) injects a dual-source principle set comprising explicit domain and implicit learned principles into a generative reward model that scores every CoT step and GUI action. Since rule compliance is checked deterministically (e.g., cursor-in-bounds, widget-state change), rewards are verifiable, preventing silent exploitation. Second, Online VM-Grounded Trajectory Construction (OVTC) spins up lightweight instances seeded with procedural task graphs; milestones declared by the agent trigger dense, step-wise rewards, enabling rapid collection of millions of high-fidelity traces without manual labeling. For example, when automating “Export chart as PDF” in LibreOffice, OVTC grants intermediate reward as soon as the agent reaches File → Export → Save action, offering fine-grained feedback unattainable in prior pipelines.

\begin{itemize}
\item We introduce Orcust, the first GUI-agent framework to integrate principle-guided reward mechanisms with generative critiques for reliable and interpretable feedback.
\item We design an automated VM-based trajectory generator that produces millions of subgoal-annotated interactions for dense, stepwise training supervision.
\item Extensive evaluations demonstrate that Orcust achieves state-of-the-art (SOTA) performance across a broad spectrum of GUI-agent tasks, including perceptual, foundational and complex multi-step scenario.
\end{itemize}
\section{Related Work}

\subsection{GUI Agent}
Early approaches~\cite{zheng2024gpt4v-webagent, gurreal2024realworld-webagent} convert GUI structures into text or DOM representations for transformer models to select elements or generate API calls. Recently, GUI agents have progressed from models depending on HTML or accessibility trees toward fully vision-based systems that perform pixel-level operations ~\cite{qin2025ui-tars, wu2024os-atlas, wan2024omniparser, hu2024dawn}. UGround~\cite{gou2024uground} trains on large-scale synthetic web data to map natural-language region descriptions to screen coordinates, and Ponder \& Press ~\cite{wang2024ponder} decouples high-level instruction interpretation from element localization, improving both modularity and accuracy. UI-R1~\cite{lu2025ui-r1} first demonstrates the DeepSeek-R1 style reinforcement learning can improve the generalization of models across diverse platforms with minimal data, while GUI-R1~\cite{xia2025gui-r1} further generalizes this paradigm by defining a unified action space across platforms with verifiable rewards for type, location, and text.

\subsection{Reinforcement Fine-Tuning}
Reward-centric post-training approaches such as DeepSeek-R1~\cite{guo2025deepseek-r1} demonstrate that rule-based rewards can incentivize reasoning in LLMs without human preference labels. subsequent studies~\cite{peng2025lmm-r1, shen2025vlm-r1, wang2025visualprm} extended RFT to multimodal Scenario by designing task-specific rewards for vision tasks. Visual-RFT~\cite{liu2025visual-rft} and Vision-R1~\cite{huang2025vision-r1} use environment-verifiable signals (e.g., IoU for image grounding and object detection) and a staged GRPO schedule to refine CoT reasoning, and R1-v~\cite{chen2025r1v} reinforce code and reasoning generation through symbolic feedback. Nevertheless, applying RFT to high-level GUI agent scenarios remains challenging due to diverse UI layouts, implicit task semantics and long-horizon action dependencies, prior RFT methods lacked advanced guidance in the reward function and did not incorporate dynamic trajectory generation. Notably, we first introduce a PCRM strategy within an RFT framework, which enables the agent to learn under explicit domain principles, and to leverage online VM-grounded trajectory construction for more scalable GUI policy learning.
\section{Method}
\begin{figure*}[htbp]
  \centering
  \includegraphics[width=1.0\linewidth]{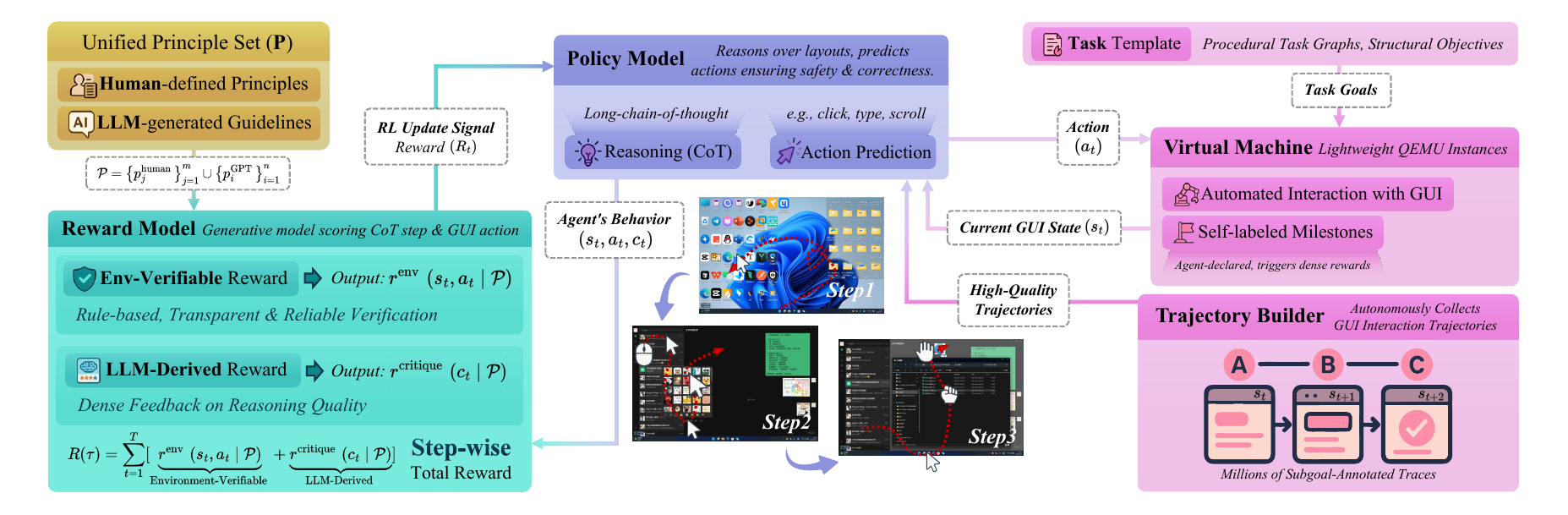}
  \caption{Architectural overview of the Orcust framework, which integrates Principle-Constrained Reward Modeling (PCRM) for stepwise, principle-aligned reward signals and Online VM-Grounded Trajectory Construction (OVTC) for automatic generation of high-quality, multi-step GUI interaction trajectories using a virtual machine environment.}
  \label{fig:main}
\end{figure*}

This section introduces two core components of Orcust: Principle-Constrained Reward Modeling (PCRM) and Online VM-Grounded Trajectory Construction (OVTC). Figure~\ref{fig:main} illustrates the framework of Orcust.

\subsection{Principle-Constrained Reward Modeling}
Inspired by the rule-based signals of multi-modal reinforcement framework~\cite{peng2025lmm-r1, chen2025r1v}, PCRM integrates deterministic validators with generative critiques to yield verifiable, interpretable rewards at each interaction step.

\paragraph{Dual-Source Principle Generation.} We derive the set of guiding principles $\mathcal{P}$ from two complementary sources: (1) Explicit domain principles, drawn from developer-provided rules and established user interface guidelines; and (2) Implicit learned principles, induced from data and language model insights. The explicit principles encapsulate known requirements and prohibitions in GUI tasks. For example, “A deletion action must be confirmed via a prompt” or “Fill all mandatory fields before submitting a form.”

In parallel, the implicit principles are extracted in a data-driven manner. We leverage large language models (LLMs) to analyze task descriptions and expert demonstrations, prompting the LLM to articulate additional plausible rules or strategies for accomplishing the tasks. For instance, given a high-level instruction and a successful demonstration, an LLM might infer a principle like “If an operation involves multiple steps (e.g., open a menu then click an option), maintain the correct sequence without skipping steps.” The outputs from both sources are consolidated into a unified principle set $\mathcal{P} = {p_1, p_2, \dots, p_K}$, where each $p_i$ is represented as a check that can be evaluated against the agent’s behavior. We implement each principle $p_i$ as a deterministic function or predicate that inspects the relevant state-action context. This function returns a boolean value $p_i(s_t, a_t)$ indicating whether the action $a_t$ taken in state $s_t$ satisfies the principle. (Some principles apply only to specific actions or situations—e.g. a confirmation rule applies only to “delete” actions—so $p_i(s_t, a_t)$ may be defined to always return True when the principle is not applicable.) By combining human knowledge with LLM-induced rules, this dual-source approach yields a robust and diverse principle set, covering both obvious constraints and more subtle high-level strategies. The principle $\mathcal{P}$ can be denoted as:

\[
\mathcal{P} = \left\{ p^{\text{human}}_j \right\}_{j=1}^{m} \cup \left\{ p^{\text{GPT}}_i \right\}_{i=1}^{n} .
\]

\paragraph{Stepwise Verifiable Rewards.} Formally, we represent an interaction trajectory as $\tau=\left\{\left(s_t, a_t, c_t\right)\right\}_{t=1}^T$, where $s_t$ is the state of the GUI environment at time step $t$ (e.g. the current screen or DOM state), $a_t$ is the action taken by the agent (e.g. a GUI operation like clicking or typing), and $c_t$ is the agent’s reasoning trace (the chain-of-thought textual context) at that step. 

Given the principle set $\mathcal{P}$, we define a reward function that evaluates the agent’s behavior at each time step by verifying principle compliance. Inspired by the process supervision methods that judge each reasoning step~\cite{lightman2023step-by-step, wang2024math-shepherd}, our reward modeling provides stepwise feedback, guiding the agent through long tasks, rather than use standard sparse rewards that only signal final success or failure. At time step $t$, the total reward for a trajectory $R(\tau)$ is computed as the sum of two components: (1) Environment-verifiable principle (EVP) reward ${r^{\text{env}}(s_t, a_t \mid \mathcal{P})}$ is determined by deterministic checks that verify the correctness of the agent's actions. Examples include ensuring that the cursor remains within bounds, the clicked widget is visible and interactive, and no crashes or unintended behaviors occur during interaction. These checks offer rule-based verification, making this reward component both transparent and reliable. (2) LLM-Derived Principle (LDP) reward $r^{\text{critique}}(c_t \mid \mathcal{P})$ provides a complementary, dense feedback signal by evaluating the quality of the agent’s reasoning and action against the principle set $\mathcal{P}$. We deploy a generative reward model $r_\theta$, which takes as input the current principles and the agent’s thought process for the step, and outputs a critique feedback and a numerical score (ranging from 1 to 10) for each reasoning step $c_t$. The model scores the agent’s adherence to the pre-defined principles, such as whether the agent is following the semantic search principle or any other relevant guideline. This mechanism resembles constitutional AI-style self-critique, where an LLM judges outputs by ethical or logical principles~\cite{bai2022constitutional-ai, sharma2025constitutional-classifier, liu2025spct}. For instance, if the agent reasons “Proceeding to delete the file without checking for save”, violating the principle “ensure no unsaved changes before deletion”, the critique model flags this and assigns a negative reward. Conversely, if the reasoning is thorough and compliant with all principles, the critique model assigns a positive reward. The result function can be defined as:

\[
R(\tau) = \sum_{t=1}^{T} \left[ \underbrace{r^{\text{env}}(s_t, a_t \mid \mathcal{P})}_{\text{Environment-Verifiable}} + \underbrace{r^{\text{critique}}(c_t \mid \mathcal{P})}_{\text{LLM-Derived}} \right].
\]

We optimize the policy via GRPO, which maximizes the advantage of each trajectory:

\[
A(\tau)=R(\tau)-\widehat{b}(group),
\]

where $\widehat{b}(group)$ represents the group-specific baseline for stability. This two-term reward ensures that every action is both verifiable (via $r^{\text{env}}$) and interpretable (via $r^{\text{critique}}$), and that reward-hacking is constrained by immutable rule checks.

\subsection{Online VM-Grounded Trajectory Construction}
Collecting high-fidelity trajectories through simulation is critical for data-efficient policy learning, especially in interactive GUI and web domains where real-world labeled sequences are scarce~\cite{qin2025ui-tars, bai2024digirl}. OVTC is the mechanism for automatically generating diverse, high-quality trajectories complete with intermediate reward annotations, without relying on manual labeling.

\paragraph{Virtual-Machine Harness.} As illustrated in Figure~\ref{fig:main}, we build a custom VM environment using lightweight QEMU/KVM instances to simulate various GUI applications and websites in a controlled setting, which are instrumented with an event bus to record every interaction with the GUI, including: (1) \textbf{Screen RGB} captures the visual state of the GUI at every time step, providing the agent with the most up-to-date image of the interface. (2) \textbf{Input event} logs every user input (e.g., mouse clicks, text entries) so the agent can correlate actions with their effects on the GUI. (3) \textbf{DOM snapshot} records the structure of the webpage or application and underlying elements of the interface, such as buttons, fields and containers, ensuring precise verification of agent actions against the corresponding DOM information.

Each task in the VM is defined by a task template which specifies a high-level goal and a sequence of requisite sub-tasks. Each task within the VM is defined by a task template that specifies a high-level objective along with its corresponding subtasks and success criteria. For example, the task "export document as PDF in a text editor" decomposes into: (1) open the "File" menu (menu DOM node becomes open and visible); (2) click "Export" and select PDF (file selection dialog appears or PDF file is created in the file system); (3) confirm the save dialog (confirmation message is displayed). These criteria are akin to environment-based checks that can be done automatically by monitoring the DOM or system events. By using a range of task templates, we cover a variety of applications (file editors, email clients, web forms, etc.) and user goals.

\subsection{Self-Labeled Sub-Goals.} GUI tasks, especially high-level ones, often require a sequence of several low-level actions to complete. To tackle the challenge of long-horizon credit assignment, we incorporate self-labeled sub-goals into the training process. For example, “Book a flight ticket” might involve sub-tasks like logging in, entering passenger info, selecting a flight, and confirming payment. Our agent auto-generates its own sub-goal markers within its reasoning process. We implement this by allowing the agent’s chain-of-thought (CoT) to emit special milestone tokens whenever it believes it has completed a meaningful intermediate objective. Concretely, we augment the agent’s output format so that it can produce a token like \texttt{[MILESTONE: ...]} (or a similar notation) in the \texttt{<think>} reasoning stream, describing the accomplishment. For instance, after the agent fills out a form in a GUI, it might insert \texttt{[MILESTONE: FormFilled]} in its reasoning trace before proceeding to the next step. These self-labeled sub-goals serve two purposes: (1) they let the agent reflect on its progress and (2) they enable the training system to automatically detect sub-goal completion and assign appropriate rewards. Upon detecting a milestone, the environment flags it in the agent’s observation and the reward model assigns a positive reward, or applies a penalty if a step was missed or incorrect.

\section{Experiments}

\subsection{Experimental Setup}
\paragraph{Training and Inference Procedures. } We instantiate Orcust in 3B and 7B parameters using a Qwen2.5-VL~\cite{bai2025qwen2.5vl} backbone. To initialize the policy and avoid instability, we first perform a brief SFT on a small curated dataset of high-quality GUI interaction trajectories, and then conduct RFT using our PCRM reward. The OVTC procedure continually generates 15K interaction trajectories by deploying the agent in a virtual machine environment, so the model can practice tasks in a realistic GUI setting.

\setlength{\tabcolsep}{1.3mm}
\renewcommand{\arraystretch}{1.08}

\begin{table*}[t]
  \centering
  \caption{GUI-grounding accuracy (\%) on \textsc{ScreenSpot-Pro} and \textsc{ScreenSpot}.  
           All models share the same zero-shot prompt for fairness.  
           * indicates supervised fine-tuning on \textsc{OVTC-15K}.}
  \footnotesize
  \begin{tabular}{
      l
      |*{12}{S}@{\hspace{0.5ex}}
      |*{4}{S}
    }
    \toprule
    \multirow{3}{*}{\textbf{Model}}
    & \multicolumn{12}{c|}{\textsc{ScreenSpot-Pro}} & \multicolumn{4}{c}{\textsc{ScreenSpot}}\\
    \cmidrule(lr){2-13} \cmidrule(lr){14-17}
    & \multicolumn{2}{c}{Dev} & \multicolumn{2}{c}{Creative} & \multicolumn{2}{c}{CAD}
    & \multicolumn{2}{c}{Scientific} & \multicolumn{2}{c}{Office} & \multicolumn{2}{c|}{OS}
    & \multicolumn{2}{c}{Web} & \multicolumn{2}{c}{Desktop}\\
    & \multicolumn{1}{c}{Text} & \multicolumn{1}{c}{Icon}
    & \multicolumn{1}{c}{Text} & \multicolumn{1}{c}{Icon}
    & \multicolumn{1}{c}{Text} & \multicolumn{1}{c}{Icon}
    & \multicolumn{1}{c}{Text} & \multicolumn{1}{c}{Icon}
    & \multicolumn{1}{c}{Text} & \multicolumn{1}{c}{Icon}
    & \multicolumn{1}{c}{Text} & \multicolumn{1}{c|}{Icon}
    & \multicolumn{1}{c}{Text} & \multicolumn{1}{c}{Icon}
    & \multicolumn{1}{c}{Text} & \multicolumn{1}{c}{Icon}\\
    \midrule
    \rowcolor[HTML]{F5F5F5}
    \multicolumn{17}{l}{\textit{Zero-shot}}\\   
    \midrule
    Qwen-VL-7B      & 0.0 & 0.0 & 0.0 & 0.0 & 0.0 & 0.0 & 0.7 & 0.0 & 0.0 & 0.0 & 0.0 & 0.0 & {--} & {--} & {--} & {--}\\
    GPT-4o          & 1.3 & 0.0 & 1.0 & 0.0 & 2.0 & 0.0 & 2.1 & 0.0 & 1.1 & 0.0 & 0.0 & 0.0 &  {--} & {--} & {--} & {--}\\
    Qwen2.5-VL-3B   & 14.9 & 2.1 & 20.2 & 1.4 & 4.1  & 4.7 & 34.0 & 7.3 & 22.0 & 3.8 & 6.5 & 2.2 & 60.0 & 43.2 & 80.9 & 40.0\\
    Qwen2.5-VL-7B   & 28.6 & 2.1 & 22.3 & 2.8 & 11.2 & 4.7 & 35.9 & 7.3 & 32.4 & 5.7 & 25.6 & 5.6 & 70.9 & 51.7 & 84.6 & 48.2\\
    \midrule
    \rowcolor[HTML]{F5F5F5}
    \multicolumn{17}{l}{\textit{Supervised fine-tuning}}\\
    \midrule
    SeeClick           & 0.6 & 0.0 & 1.0 & 0.0 & 2.5 & 0.0 & 3.5 & 0.0 & 1.1 & 0.0 & 2.8 & 0.0 & 55.7 & 32.5 & 72.2 & 30.0\\
    Os-Atlas-4B        & 7.1 & 0.0 & 3.0 & 1.4 & 2.0 & 0.0 & 9.0 & 5.5 & 5.1 & 3.8 & 5.6 & 0.0 & 82.6 & 63.1 & 72.1 & 45.7\\
    Os-Atlas-7B        & 33.1 & 1.4 & 28.8 & 2.8 & 12.2 & 4.7 & 37.5 & 7.3 & 33.9 & 5.7 & 27.1 & 4.5 & 90.8 & 74.2 & 91.7 & 62.8\\
    ShowUI-2B          & 16.9 & 1.4 & 9.1 & 0.0 & 2.5 & 0.0 & 13.2 & 7.3 & 15.3 & 7.5 & 10.3 & 2.2 & {--} & {--} & {--} & {--}\\
    CogAgent-18B       & 14.9 & 0.7 & 9.6 & 0.0 & 7.1 & 3.1 & 22.2 & 1.8 & 13.0 & 0.0 & 5.6 & 0.0 & 70.4 & 28.6 & 74.2 & 20.0\\
    Aria-GUI           & 16.2 & 0.0 & 23.7 & 2.1 & 7.6 & 1.6 & 27.1 & 6.4 & 20.3 & 1.9 & 4.7 & 0.0 & {--} & {--} & {--} & {--}\\
    UGround-7B         & 26.6 & 2.1 & 27.3 & 2.8 & 14.2 & 1.6 & 31.9 & 2.7 & 31.6 &11.3 & 17.8 & 0.0 & 80.4 & 70.4 & 82.5 & 63.6\\
    Qwen2.5-VL-3B$^{*}$& 18.7 & 2.1 & 25.1 & 2.8 & 12.3 & 5.1 & 38.6 & 6.8 & 29.2 & 5.7 & 18.3 & 2.2 & 64.4 & 46.3 & 84.1 & 48.5\\
    Qwen2.5-VL-7B$^{*}$& 33.0 & 2.8 & 28.8 & 3.5 & 14.9 & 4.7 & 42.6 & 7.3 & 38.2 & 8.9 & 31.9 & 5.6 & 77.9 & 57.4 & 87.0 & 58.5\\
    \midrule
    \rowcolor[HTML]{F5F5F5}
    \multicolumn{17}{l}{\textit{Reinforcement fine-tuning}}\\
    \midrule
    UI-R1-3B   & 22.7 & 4.1 & 27.3 & 3.5 & 11.2 & 6.3 & 43.4 & 11.8 & 32.2 & 11.3 & 13.1 & 4.5 & 85.2 & 73.3 & 90.2 & 59.3\\
    GUI-R1-3B  & 33.8 & 4.8 & 40.9 & 5.6 & 26.4 & 7.8 & 61.8 & 17.3 & 53.6 & 17.0 & 28.1 & 5.6 & 89.6 & 72.1 & 93.8 & 64.8\\
    GUI-R1-7B  & 49.4 & 4.8 & 38.9 & 8.4 & 23.9 & 6.3 & 55.6 & 11.8 & 58.7 & 26.4 & 42.1 & 16.9 & 91.3 & 75.7 & 91.8 & 73.6\\
    \rowcolor[HTML]{FFF9E6}
    \textbf{Orcust-3B}
               & 48.6 & 7.0 & 49.2 & 9.1 & \bfseries37.2 & 9.4 & 65.2 & 19.1 & 62.7 & 26.4 & 40.4 & 12.4 & 93.6 & 77.2 & \bfseries97.2 & 72.2\\
    \rowcolor[HTML]{FFF9E6}
    \textbf{Orcust-7B}
               & \bfseries60.4 & \bfseries12.4 & \bfseries58.3 & \bfseries12.2 & 35.8 & \bfseries14.1 & \bfseries69.5 & \bfseries24.9 & \bfseries73.3 & \bfseries32.3 & \bfseries56.6 & \bfseries18.2 & \bfseries95.8 & \bfseries83.7 & 95.8 & \bfseries79.8\\
    \bottomrule
  \end{tabular}
  \label{tab:grounding}
\end{table*}

\paragraph{Evaluation Benchmarks. } We evaluate Orcust on eight standard GUI agent benchmarks, covering mobile, desktop and web environments. Low-level benchmarks include AndroidControl-Low~\cite{li2024android-control} (mobile UI step-by-step tasks), GUI-Act-Web~\cite{chen2024gui-act}, OmniAct-Web and OmniAct-Desktop~\cite{kapoor2024omniact} (web platform and desktop environment actions). High-level, multi-step benchmarks include AndroidControl-High~\cite{li2024android-control} (mobile long-horizon tasks) and GUI-Odyssey~\cite{lu2024gui-odyssey} (cross-application navigation scenarios). For assessing visual grounding capabilities, we use the ScreenSpot~\cite{cheng2024seeclick} benchmark that spans GUI understanding tasks on mobile/desktop/web, and the ScreenSpot-Pro~\cite{li2025screenspot-pro}, featuring high-resolution professional UIs with expert-annotated targets, covering 23 apps across 5 industries and 3 operating systems. 

\paragraph{Evaluation Metrics. } Following Os-Atlas~\cite{wu2024os-atlas}, we use the standard metrics for GUI agent evaluation, including Type accuracy (exact match of the predicted action type, e.g. Click vs Scroll), Grounding accuracy (correctness of the predicted click position or target UI element), and Stepwise Success Rate (a step is successful only if both the action type and all associated arguments, e.g. click coordinates or input text are correct).

\subsection{Main Result}
We compare Orcust-3B/7B with prior SOTA models, including the reinforcement-learned UI-R1~\cite{lu2025ui-r1} and GUI-R1~\cite{xia2025gui-r1}, as well as strong supervised models UGround~\cite{gou2024uground}, OS-Atlas~\cite{wu2024os-atlas}.

\setlength{\tabcolsep}{2mm}  
\renewcommand{\arraystretch}{1.15}  

\begin{table*}[t]
\centering
\caption{GUI low-level task accuracy on \textsc{AndroidControl-Low}, \textsc{GUI-Act-Web}, \textsc{OmniAct-Web} and \textsc{OmniAct-Desktop}. All experiments are conducted under the same zero-shot prompt for fair comparison.}
\resizebox{\linewidth}{!}{
\begin{tabular}{
    l
    |>{\centering\arraybackslash}p{1cm} >{\centering\arraybackslash}p{1cm} >{\centering\arraybackslash}p{1cm} 
    |>{\centering\arraybackslash}p{1cm} >{\centering\arraybackslash}p{1cm} >{\centering\arraybackslash}p{1cm} 
    |>{\centering\arraybackslash}p{1cm} >{\centering\arraybackslash}p{1cm} >{\centering\arraybackslash}p{1cm} 
    |>{\centering\arraybackslash}p{1.1cm} >{\centering\arraybackslash}p{1.1cm} >{\centering\arraybackslash}p{1.1cm} 
    |c
}
\toprule
\multirow{2}{*}{\textbf{Models}} & \multicolumn{3}{c|}{\textsc{GUI-Act-Web}} & \multicolumn{3}{c|}{\textsc{OmniAct-Web}} & \multicolumn{3}{c|}{\textsc{OmniAct-Desktop}} & \multicolumn{3}{c|}{\textsc{AndroidControl-Low}} & \multirow{2}{*}{\textbf{Overall}}  \\
& \textbf{Type} & \textbf{GR} & \textbf{SR} & \textbf{Type} & \textbf{GR} & \textbf{SR} & \textbf{Type} & \textbf{GR} & \textbf{SR} & \textbf{Type} & \textbf{GR} & \textbf{SR} & \\
\midrule
\rowcolor[HTML]{F7F7F7}
\multicolumn{14}{l}{\textit{Supervised Fine-Tuning}}\\
\midrule
Os-Atlas-4B       & 79.22 & 58.57 & 42.62 & 46.74 & 49.24 & 22.99 & 63.30 & 42.55 & 26.94 & 64.58 & 71.19 & 40.62 & 50.71 \\
Os-Atlas-7B       & 86.95 & 75.61 & 57.02 & 85.63 & 69.35 & 59.15 & 90.24 & 62.87 & 56.73 & 73.00 & 73.37 & 50.94 & 70.07 \\
\midrule
\rowcolor[HTML]{F7F7F7}
\multicolumn{14}{l}{\textit{Reinforcement Fine-Tuning}}\\
\midrule
UI-R1-3B          & 75.89 & 79.43 & 67.31 & 75.42 & 61.35 & 61.33 & 73.41 & 64.12 & 63.98 & 79.15 & 82.41 & 66.44 & 70.85 \\
GUI-R1-3B         & 89.86 & 87.42 & 76.31 & 88.58 & 75.10 & 75.08 & 91.86 & 78.37 & 78.31 & 83.68 & 81.59 & 64.41 & 80.88 \\
GUI-R1-7B         & 90.85 & 88.06 & 80.31 & 91.16 & 77.29 & 77.35 & 92.20 & 83.36 & 83.33 & 85.17 & 84.02 & 66.52 & 83.30 \\
\rowcolor[HTML]{FFF9E6}
\textbf{Orcust-3B} & 94.23 & \textbf{93.14} & 83.22 & \textbf{94.77} & 79.15 & 81.63 & 95.88 & 84.55 & 84.73 & 86.32 & 84.94 & 68.19 & 85.90 \\
\rowcolor[HTML]{FFF9E6}
\textbf{Orcust-7B} & \textbf{96.41} & 92.44 & \textbf{87.76} & \textbf{94.24} & \textbf{81.83} & \textbf{83.25} & \textbf{95.15} & \textbf{87.94} & \textbf{88.02} & \textbf{89.75} & \textbf{87.85} & \textbf{72.32} & \textbf{88.08} \\
\bottomrule
\end{tabular}}
\label{tab:low_level}
\end{table*}

\setlength{\tabcolsep}{1.5mm}  
\renewcommand{\arraystretch}{1.1}  

\begin{table}[t]  
\centering
\caption{GUI high-level task accuracy on \textsc{GUI-Odyssey} and \textsc{AndroidControl-High}. * denotes supervised fine-tuned on \textsc{OVTC-15K}.}
\resizebox{1.0\linewidth}{!}{  
\begin{tabular}{
    l
    |>{\centering\arraybackslash}p{1.2cm} >{\centering\arraybackslash}p{1.2cm} >{\centering\arraybackslash}p{1.2cm} 
    |>{\centering\arraybackslash}p{1cm} >{\centering\arraybackslash}p{1cm} >{\centering\arraybackslash}p{1cm}
}
\toprule
\multirow{2}{*}{\textbf{Models}} & \multicolumn{3}{c|}{\textsc{AndroidControl-High}} & \multicolumn{3}{c}{\textsc{GUI-Odyssey}} \\
& \textbf{Type} & \textbf{GR} & \textbf{SR} & \textbf{Type} & \textbf{GR} & \textbf{SR} \\
\midrule
\rowcolor[HTML]{F7F7F7}
\multicolumn{7}{l}{\textit{Zero Shot}}\\
\midrule
GPT-4o            & 63.06 & 30.90 & 21.17 & 37.50 & 14.17 & 5.36  \\
Qwen2.5-VL-3B     & 45.66 & 47.23 & 36.18 & 35.10 & 27.15 & 28.06 \\
Qwen2.5-VL-7B     & 59.70 & 55.35 & 45.13 & 49.20 & 38.29 & 33.85 \\
\midrule
\rowcolor[HTML]{F7F7F7}
\multicolumn{7}{l}{\textit{Supervised Fine-Tuning}}\\
\midrule
OS-Atlas-4B & 49.01 & 49.51 & 22.77 & 49.63 & 34.63 & 20.25  \\
OS-Atlas-7B &  57.44 & 54.90 & 29.83 & 60.42 & 39.74 & 26.96  \\
QwenVL2.5-3B* & 50.17 & 47.55 & 37.29 & 37.18 & 28.33 & 27.25  \\
QwenVL2.5-7B* & 58.32 & 56.17 & 44.18 & 50.54 & 37.72 & 35.10 \\
\midrule
\rowcolor[HTML]{F7F7F7}
\multicolumn{7}{l}{\textit{Reinforcement Fine-Tuning}}\\
\midrule
UI-R1-3B         & 57.85 & 55.70 & 45.44 &  52.16 & 34.46 & 32.49 \\
GUI-R1-3B        & 58.04 & 56.24 & 46.55 & 54.84 & 41.52 & 41.33  \\
GUI-R1-7B        & 71.63 & 65.56 & 51.67 &  65.49 & 43.64 & 38.79  \\
\rowcolor[HTML]{FFF9E6}
\textbf{Orcust-3B} & 72.55 & 66.18 & 57.29 & 69.12 & \textbf{54.15} & 51.15 \\
\rowcolor[HTML]{FFF9E6}
\textbf{Orcust-7B} & \textbf{79.12} & \textbf{79.42} & \textbf{62.93} & \textbf{75.25} & 53.88 & \textbf{56.34} \\
\bottomrule
\end{tabular}}
\label{tab:high_level}
\end{table}

\paragraph{Grounding Capability. } We summarize the GUI grounding accuracy results on the ScreenSpot and ScreenSpot-Pro benchmarks in Table~\ref{tab:grounding}. Orcust exhibits SOTA grounding performance, significantly outperforming all baselines in both text and icon grounding. Orcust-3B achieves higher average accuracy than the previous SOTA model OS-Atlas-7B, underscoring the data-efficiency of our principle-aligned RL approach (15K trajectories versus 14M SFT samples). Notably, Orcust-7B attains an average of 88.8\% accuracy on the ScreenSpot (Web + Desktop) tasks and 39.0\% on text and icon categories across six ScreenSpot-Pro domains, about 5.7\% and 13.8\% higher than the GUI-R1-7B. We attribute these gains to the principle-based rewards guiding the model to focus on correct high-level grounding, allowing Orcust to leverage small but diverse data for generalizable GUI understanding. We also observe that simply fine-tuning the base model on high-quality data can improve grounding, for instance, Qwen2.5-VL-3B/7B* supervised on the OVTC-15K dataset outperforms its zero-shot counterparts by roughly 4.8\% and 6.4\% points on ScreenSpot. Nonetheless, the principle-constrained reward modeling leads to far greater gains, guiding the agent to align UI elements with instructions more effectively.

\paragraph{Low-level Task Capability. } We evaluate low-level action execution on the four step-by-step benchmarks (GUI-Act-Web, OmniAct-Web, OmniAct-Desktop, AndroidControl-Low). As shown in Table~\ref{tab:low_level},  Orcust-7B delivers the best performance on every metric, achieving 96.4\% action type accuracy, 92.4\% grounding accuracy, 82.8\% step-wise accuracy, and an overall success rate of 88.1\%, substantially outperforming the previous best 7B model (GUI-R1-7B at 83.3\%) and the strongest supervised baseline (OS-Atlas at 70.1\%). Even the smaller Orcust-3B model (85.9\% success) surpasses the GUI-R1-7B, underscoring the efficiency of our principle-constrained RL approach. We observe that Orcust mitigates the overfitting issues seen in small-scale SFT, which even degrade certain low-level metrics (e.g. grounding accuracy on AndroidControl-Low), whereas Orcust’s policy remains robust on single-step action execution across mobile, web, and desktop tasks.

\paragraph{High-level Task Capability. } As illustrated in Table~\ref{tab:high_level}, Orcust pronounce significant advantages across high-level task execution on AndroidControl-High and GUI-Odyssey, which involve multi-step tasks and cross-application navigation. Orcust-3B outperforms the previously proposed RL methods UI-R1-3B and GUI-R1-3B by approximately 14.39\% and 10.99\% respectively, while Orcust-7B achieves an average score of 67.80\%, significantly surpassing all baseline models. Furthermore, we observe that Qwen2.5-VL exhibits marginal performance improvement after supervised fine-tuning. It demonstrate the critical impact of PCRM in long-horizon reasoning. By encoding high-level GUI principles into the reward function, PCRM guides the agent to plan more coherently and avoid faulty action sequences, leading to markedly higher goal-completion rates. We see that Orcust not only selects correct action types but also executes them with greater accuracy. Compared to earlier methods, Orcust’s trajectories have fewer failed steps and less deviation from the task goal, which boosts the overall success in complex multi-step scenarios. In short, PCRM enables more reliable action planning and proper GUI usage principles, thus maintains high success even as task length grows, a key differentiator on tasks like cross-app navigation and form-filled automation.

\subsection{Ablation Study} To understand the impact of our reward type, data curation and reward granularity in Orcust, we conduct extensive ablation experiments on Orcust-7B.

\setlength{\tabcolsep}{1.5mm}{
\begin{table}[t] 
\centering
\caption{Ablation Study in the reward function.} 
\resizebox{1\linewidth}{!}{
\begin{tabular}{l|ccc | ccc}
\toprule
\multirow{2}{*}{Reward Type}  & \multicolumn{3}{c}{AndroidControl-Low}  & \multicolumn{3}{c}{AndroidControl-High} \\
 & Type & GR & SR & Type & GR & SR   \\
\midrule
EVP-only   & 87.17 & 86.25 & 69.12 & 77.09 & 77.87 & 57.52 \\
LDP-only   & 86.82 & 85.42 & 70.31 & 77.17 & 77.65 & 60.23 \\
\rowcolor[HTML]{FFF9E6}
\textbf{EVP \& LDP} & \bfseries89.75 & \bfseries87.85 & \bfseries72.32 & \bfseries79.12 & \bfseries79.42 & \bfseries62.93  \\
\bottomrule
\end{tabular}}
\label{tab:abl_rft}
\end{table} 
}

\setlength{\tabcolsep}{1.5mm}{
\begin{table}[t] 
\centering
\caption{Ablation Study in the average reward step.} 
\resizebox{1\linewidth}{!}{
\begin{tabular}{l|ccc | ccc}
\toprule
\multirow{2}{*}{Reward Step}  & \multicolumn{3}{c}{AndroidControl-Low}  & \multicolumn{3}{c}{AndroidControl-High} \\
 & Type & GR & SR & Type & GR & SR   \\
\midrule
1-step   & 87.33 & 86.45 & 69.33 & 76.49 & 77.11 & 58.19 \\
2-step   & \bfseries90.85 & 86.41 & 71.61 & 78.36 & \bfseries80.55 & 61.32 \\
\rowcolor[HTML]{FFF9E6}
4-step   & 89.75 & \bfseries87.85 & \bfseries72.32 & \bfseries79.12 & 79.42 & \bfseries62.93  \\
8-step   & 87.42 & 86.13 & 71.89 & 77.60 & 77.95 & 61.66 \\
\bottomrule
\end{tabular}}
\label{tab:abl_rs}
\end{table} 
}

\begin{figure*}[htbp]
  \centering
  \includegraphics[width=0.95\linewidth]{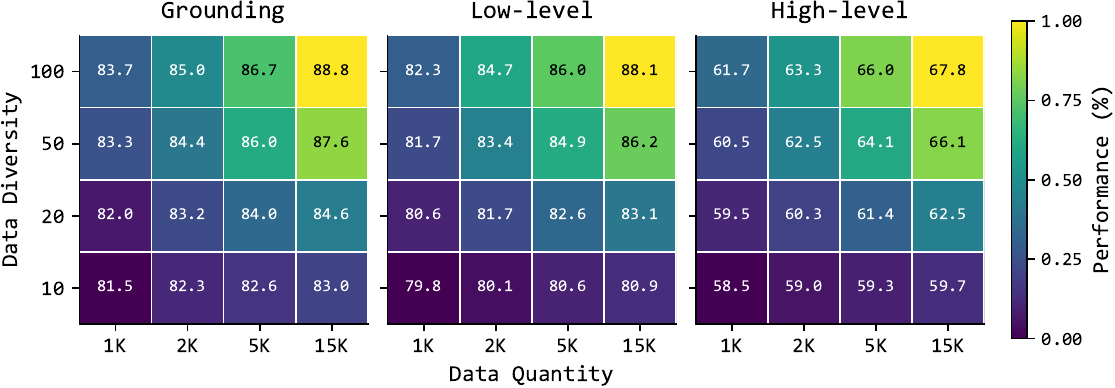}
  \caption{Performance heatmap (\%) vs. data diversity (vertical) and data quantity (horizontal).}
  \label{fig:trajectory_performance}
\end{figure*}

\begin{figure}[htbp]
  \centering
  \includegraphics[width=0.95\linewidth]{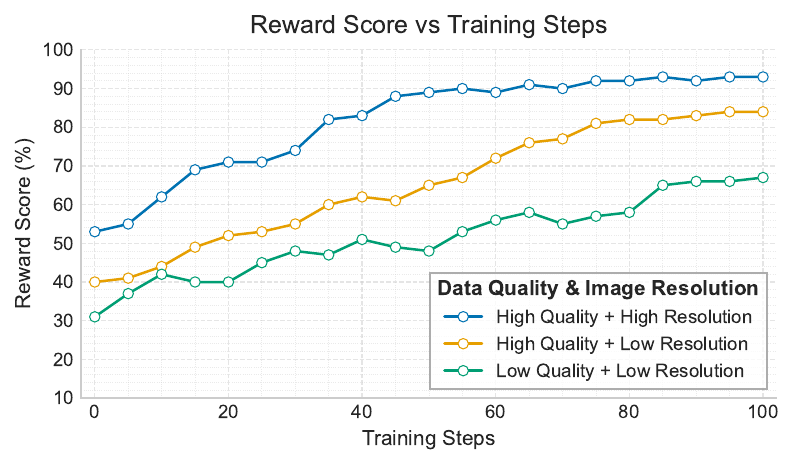}
  \caption{Reward score convergence with varying data quality and image resolution. }
  \label{fig:quality_resolution_reward}
\end{figure}

\paragraph{Reward Function Types. } We ablate the reward function to compare the contributions of Environment-Verifiable Principles (EVP) versus LLM-Derived Principles (LDP) in Table~\ref{tab:abl_rft}. The hybrid EVP\&LDP consistently achieves the average score of 78.57\% on the AndroidControl-Low and AndroidControl-High tasks, outperforming the EVP-only variant by 2.73\% and LDP-only variant by 2.30\%. Specifically, EVP covers immediate, factual correctness to prevent outright mistakes, while LDP encourages strategic coherence and adherence to the instruction’s intent. Together, the hybrid approach inherits the strengths and offers a more comprehensive feedback. In qualitative terms, the EVP-only agent tends to “click correctly” but sometimes lacks context and follows instructions too literally, whereas the LDP-only agent is more context-aware but can occasionally drift since the LLM feedback isn’t tied to guaranteed ground-truth. 

\paragraph{Stepwise Reward Depth. } We investigate how the number of interaction steps over which rewards are averaged or assigned impacts learning. As illustrated in Table~\ref{tab:abl_rs}, we observe that an intermediate feedback horizon yields the best performance, whereas too short or too long intervals underperform. In AndroidControl-Low, using a 4-step reward produces the highest SR of 72.32\%, versus 69.33\% with 1-step feedback. A similar improvement appears in the harder AndroidControl-High tasks, where the SR rises from 58.19\% at 1-step to 62.93\% at 4-step feedback It confirms that short-horizon rewards may fail to credit longer-term actions, while overly delayed reward make credit assignment difficult, slightly reducing performance. Thus, a 4-step reward interval provides the most informative feedback for policy learning.

\paragraph{Trajectory Diversity and Quantity. } In Figure~\ref{fig:trajectory_performance}, we analyze the impact of training dataset diversity (number of distinct tasks/UI contexts) and quantity (number of trajectories) on the agent’s performance at the average score of grounding (ScreenSpot), low-level and high-level tasks. We observe that under the lowest data diversity setting, increasing the data quantity from 1K to 15K results in performance improvements of only 1.5\%, 1.1\% and 1.8\% across the three tasks, respectively, whereas the improvements are 5.1\%, 5.8\% and 6.1\% under the highest data diversity setting. Moreover, the results demonstrate that diversity provides strong gains in any data regime, while insufficient data limits the performance ceiling. Therefore, both dimensions matter that an abundance of training trajectories alone cannot cover all cases if they lack diversity, and a diverse set of tasks will still leave many gaps if it’s too small. It justifies that OVTC gathers a reasonably large number of trajectories spanning many different GUI contexts, enabling Orcust to learn skills that transfer broadly.

\paragraph{Data Quality and Image Resolution. } As shown in the reward curves of Figure~\ref{fig:quality_resolution_reward}, using high-quality VM-grounded trajectories and high-resolution screenshots significantly accelerates learning and improves the final reward attainment. The High Quality + High Resolution setting achieves a high reward score within the first 20 training steps and eventually plateaus around 90\%, whereas the Low Quality + Low Resolution baseline climbs much more slowly and only reaches 60\% even after 100 training steps. It indicates higher image resolution clearly enables the agent to perceive fine UI details such as small text, tiny icons, which would be blurred or lost in low-res inputs, thus improving its grounding accuracy and confidence. Likewise, filtering out low-quality trajectories (e.g. those with wrong labels or irrelevant steps) accelerates learning.
\section{Conclusion}
In this work, we present Orcust, a novel framework for GUI agents that seamlessly integrates Principle-Constrained Reward Modeling with Online VM-Grounded Trajectory Construction. PCRM combines verifiable, rule-based rewards with LLM-derived critiques to provide stepwise, interpretable feedback, while OVTC harnesses lightweight VM to autonomously generate millions of high-fidelity interaction traces. Experiments demonstrate that Orcust achieves SOTA performance across eight GUI benchmarks, including ScreenSpot and ScreenSpot-Pro, highlighting that the integration principle-based reward constraints with scalable data collection significantly enhances robustness and data efficiency in interactive GUI tasks.

\section*{Limitations}
Despite these promising results, our approach has several limitations. First, Orcust relies on simulated VM environments for training and evaluation. While this ensures controlled and repeatable experiments, it may not capture the full variability of real-world GUIs. Then, the computational overhead of our method is significant: running VM-based simulations and computing dense stepwise rewards (including querying an LLM for critique feedback) incurs a high cost. This complexity could impede scalability or real-time use, as training and inference require substantial resources and careful optimization.

Another concern is deploying an autonomous GUI agent in practical settings raises ethical and security considerations. Its advanced automation capabilities could be misapplied to sensitive or critical GUI tasks (for example, manipulating financial or personal data through the interface) without proper authorization or oversight. There is also a privacy risk when scaling to real-world interfaces, as the agent may handle or observe user data on-screen. These issues highlight the need for careful safeguards, transparency, and possibly policy constraints before applying Orcust to real-world applications. Ensuring that the system’s actions remain ethical, user-approved, and within intended bounds will be crucial as we move from simulation to deployment.

\bibliography{acl_latex}
\clearpage

\appendix
\section{Appendix}

\subsection{Prompt for Training and Inference}
During training, the Orcust agent is guided by Principle-Constrained Reward Modeling (PCRM) prompts. These prompts integrate explicit domain principles and implicit LLM-derived rules directly into the feedback loop. After each action, the prompt includes annotated checks for each principle/rule and computes a stepwise reward. This yields a dense reward signal at every step, with contributions from deterministic rule checks and LLM-based critiques. Listing~\ref{prompt_training_example} are examples for training and inference prompt in raw text form, showing how rules are enforced and how rewards are scored at each step.

During reinforcement fine-tuning, each turn in an episode is formatted as a single concatenated prompt. This prompt is designed to provide the agent with all necessary context and to structure its output for effective learning. The prompt includes: (i) The high-level task description. (ii) The current GUI screenshot, representing the state $s_t$. (iii) An instruction stub that enforces the \texttt{<think>-<answer>} structure.

This \texttt{<think>-<answer>} structure encourages the agent to first output its chain-of-thought (CoT) reasoning within the \texttt{<think>} tags. This reasoning can include its step-by-step plan, assessment of the current GUI state, and importantly, self-identified sub-goals denoted by \texttt{[MILESTONE: ...]} tokens. Following the reasoning, the agent provides the GUI action in JSON format within the \texttt{<answer>} tags. This structured output is crucial for PCRM, which evaluates both the reasoning and the action to provide stepwise, interpretable rewards.

\vspace{\baselineskip}
\noindent \textbf{\textcolor{red}{Explanation of Key Elements.}}
\begin{enumerate}[label=\arabic*., wide, labelindent=0pt]
    \item \textbf{Stepwise Input:} The prompt provides the current state ($s_t$ via screenshot) and task context at each step $t$.

    \item \textbf{CoT Generation (\texttt{<think>} block):}
    \begin{itemize}[label=\textbullet]
        \item The agent generates its reasoning $c_t$ here. This is crucial for the \textbf{LLM-Derived Principle (LDP)} reward component of PCRM. The generative reward model ($r_c$) evaluates this $c_t$ against the principle set $\mathcal{P}$.
        \item \textbf{Self-Labeled Sub-Goals}: The agent can emit \texttt{[MILESTONE: FormFilled]} or \texttt{[MILESTONE: NavigatedToExportMenu]} tokens within this block. The OVTC environment detects these, and the reward model assigns a positive reward (or penalty if incorrect/missed), contributing to dense, step-wise rewards.
    \end{itemize}

    \item \textbf{Action Generation (\texttt{<answer>} block):}
    \begin{itemize}[label=\textbullet]
        \item The agent predicts an atomic action $a_t$. This action is primarily evaluated by the \textbf{Environment-Verifiable Principle (EVP)} reward component of PCRM. Checks include cursor-in-bounds, widget visibility/interactivity, etc.
        
        \item The JSON structure ensures the action is parseable and directly executable in the VM environment.
    \end{itemize}

    \item \textbf{Online VM-Grounded Trajectory Construction (OVTC):} This prompt is used by the agent deployed within the VM. The agent's \texttt{<answer>} is executed in the VM, leading to a new state $s_{t+1}$, and the cycle continues, generating trajectories for RFT.
\end{enumerate}

\vspace{\baselineskip}
\noindent
\textbf{\textcolor{red}{Placeholders.}}
\begin{itemize}[nosep,leftmargin=1.5em]
  \item \verb|{SCREENSHOT_BASE64}| - PNG of state~$s_t$ encoded as a data~URI.%
  \item \verb|type| $\in$ \{\texttt{click}, \texttt{type}, \texttt{scroll}, \texttt{key}\}.%
  \item \verb|target| - absolute \texttt{"x"}, \texttt{"y"} coordinates or a UI-tree element ID.%
  \item \verb|text| - non-empty only when \verb|type == "type"|.%
  \item \verb|t|: The current step number within the episode.

\end{itemize}

\subsection{Illustrative Input for Reward Model Critique}
The Principle-Constrained Reward Modeling (PCRM) framework evaluates the agent's behavior at each step. A core part of this is the LDP component, which uses a generative reward model ($r_c$) to critique the agent's reasoning ($c_t$) and action ($a_t$) against the defined principle set (\texttt{P}). To perform this critique, the reward model $r_c$ receives a structured input containing all relevant information for the current step. While the paper does not specify the exact prompt tokenization provided to $r_c$, we can construct an illustrative example (Listing~\ref{prompt_reward_example}) of the information it would process, designed to enable a comprehensive evaluation.

\vspace{\baselineskip}
\noindent \textbf{\textcolor{red}{Illustrative Input for Critique Model.}}
\begin{enumerate}[label=\arabic*., wide, labelindent=0pt]
    \item \textbf{The High-Level Task Context:} This provides the overall goal the agent is trying to achieve, ensuring the critique model understands the broader objective.
    \item \textbf{The Current GUI State ($s_t$):} The visual screenshot of the interface at time $t$, allowing the critique model to ground the agent's reasoning and action in the actual UI.
    \item \textbf{The Agent's Behavior at Step $t$:} This is a crucial part, encompassing:
    \begin{itemize}[label=\textbullet, leftmargin=*]
        \item \textbf{Agent's Reasoning ($c_t$):} The chain-of-thought output from the agent, including any self-labeled sub-goals (\texttt{[MILESTONE]} tokens).
        \item \textbf{Agent's Predicted Action ($a_t$):} The GUI action (e.g., click, type) in JSON format that the agent proposes to take.
    \end{itemize}
    \item \textbf{Applicable Principles for Evaluation:} A relevant subset of the guiding principles (\texttt{P}), both human-defined (EVP-related) and LLM-derived (LDP-related), against which the agent's behavior is to be judged.
    \item \textbf{The Reward Model's Task Instructions:} Specific instructions to the critique model on how to evaluate the agent's behavior against the principles and what format the output critique and score should take.
\end{enumerate}

\noindent \textbf{\textcolor{red}{Explanation of Key Elements.}}
\begin{enumerate}[label=\arabic*., wide, labelindent=0pt]

    \item \textbf{Task Context and GUI State:} These elements provide the necessary grounding. The critique model needs to understand what the agent is \textit{supposed} to do and what the visual environment \textit{looks like} to assess the appropriateness of the agent's reasoning and action.

    \item \textbf{Agent's Reasoning ($c_t$):} This is the primary input for LDP evaluation. The critique model assesses:
    \begin{itemize}[label=\textbullet]
        \item \textbf{Coherence:} Does the reasoning logically lead to the proposed action?
        \item \textbf{Completeness:} Has the agent considered all necessary preconditions or safety checks (e.g., as per \texttt{LDP\_Completeness})?
        \item \textbf{Efficiency:} Is the agent avoiding redundant steps (e.g., as per \texttt{LDP\_Efficiency})?
        \item \textbf{Adherence to Sub-goals:} If \texttt{[MILESTONE]} tokens are used, does the reasoning reflect progress towards them?
    \end{itemize}

    \item \textbf{Agent's Action ($a_t$):} While EVPs directly check action verifiability, the LDP critique also considers the action in light of the reasoning. For instance, a correctly formatted action might still be flagged by LDP if the reasoning leading to it was flawed or violated a strategic principle.

    \item \textbf{Applicable Principles:} Supplying the specific principles focuses the critique model's evaluation. This allows for targeted feedback related to safety, UI state consistency, coherence, completeness, and efficiency.

    \item \textbf{Reward Model Task Instructions:} These guide the critique model to produce a structured output, typically a numerical score (e.g., 1-10 for LDPs) and a textual critique explaining the score, referencing specific principles that were met or violated. This output then contributes to the $r^{\text{critique}}$ component of the total reward.
\end{enumerate}

\subsection{Hyperparameter Setting}
We configure the hyperparameters as listed in Table~\ref{tab:hyperparameters} and train the base model using 32 NVIDIA A100 GPUs. The Orcust model is instantiated using the Qwen2.5-VL backbone. Our training regimen comprises two main phases: an initial Supervised Fine-Tuning (SFT) phase and a subsequent Reinforcement Fine-Tuning (RFT) phase. 

For the Initial SFT Phase, the model is fine-tuned for 1-3 epochs on a curated dataset of approximately 1K-5K high-quality GUI interaction trajectories. We employ the AdamW optimizer with a learning rate ranging from 2e-5 to 5e-6 and a batch size of 16-32. The maximum prompt length is set between 1024 and 2048 tokens.

In the RFT Phase, policy refinement is conducted using PCRM. The Online VM-Grounded Trajectory Construction (OVTC) module continually generates approximately 15K interaction trajectories. The learning rate for RFT is set between 1e-6 and 5e-7. The generative reward model (r\_critique) component of PCRM utilizes a Qwen2.5-72B. During RFT, the training batch size is 128-256. To manage policy divergence, a KL divergence constraint between 0.01 and 0.05 is applied, typical for GRPO-style algorithms. For online trajectory generation within OVTC, the self-labeled milestone tokens are assigned illustrative reward/penalty values of +0.5/-0.5.

\lstset{
  language=Python,
  escapeinside={(*}{*)}  
}

\begin{lstlisting}[float=*,language=,frame=single,basicstyle=\ttfamily\small,caption=The training prompt for the Orcust,label=prompt_training_example]
### SYSTEM
You are a GUI agent.
First write your reasoning between <think> ... </think>.
Then output ONE GUI action inside <answer> ... </answer> in JSON.

### Task
Export the active chart as a PDF file and close the dialog.

<principles>
P1  Cursor must remain inside the visible viewport.
P2  Destructive operations require explicit confirmation.
P3  Maintain the action sequence specified by the menu hierarchy.
</principles>

<screen>
![screenshot]({SCREENSHOT_BASE64}) @ step {t}
</screen>

<think>
[MILESTONE: MenuOpened]
I have opened **File (*$\rightarrow$*) Export** (P3 satisfied).  
Next, according to P2 I must choose "PDF" before confirming.  
Cursor (412, 265) is within bounds (P1).
</think>

<answer>
{"type":"click", "target":{"x":412,"y":265}, "text":""}
</answer>
\end{lstlisting}

\begin{lstlisting}[float=*,language=,frame=single,basicstyle=\ttfamily\small]
### SYSTEM
You are Orcust, a vision-language GUI agent.
You must follow the **Principles** below at every step.
After thinking, output a JSON dict with keys "thought" and "action".

### TASK
Create a new folder named `Work` on the desktop.

<principles>
1. Act only on visible interactive elements.
2. Decompose long tasks into explicit sub-goals.
3. Mark a **MILESTONE** token immediately after a sub-goal is really achieved.
4. Never guess-verify on screen before proceeding.
5. Minimise unnecessary clicks / keystrokes.
</principles>

<screen>
![screenshot]({SCREENSHOT_BASE64}) @ step {t}
</screen>  

FORMAT RULES:
- Enclose chain-of-thought in **<think> ... </think>**.  
- End with **<answer> ... </answer>** containing a single JSON action,  
  e.g. {"type":"click","x":512,"y":164} or {"type":"type","text":"Work"}.

<think>
Goal: create folder "Demo".  
Step1: Right-click empty area (*$\rightarrow$*) [MILESTONE: sub-menu open]  
Step2: Select "New (*$\rightarrow$*) Folder" (*$\rightarrow$*) [MILESTONE: folder shell]
Step3: Type name and press Enter (*$\rightarrow$*) [MILESTONE: folder created]
</think>

<answer>
{"type":"click","x":732,"y":645}
</answer>
\end{lstlisting}

\begin{lstlisting}[float=*,language=,frame=single,basicstyle=\ttfamily\small,breaklines=true,caption=The critique template for the generative reward model,label=prompt_reward_example]
### Task Context
Instruction: "Delete the draft proposal document, ensuring no unsaved changes are lost."
Current Screen State ((*$s_t$*)): [Image data or rich representation of the GUI where a 'Delete' button is visible, and an 'Unsaved Changes' warning is NOT visible]

### Agent's Behavior at Step t
Agent's Reasoning (c_t):
<think>
The task is to delete 'draft_proposal.doc'.
First, I need to ensure there are no unsaved changes. The UI does not show any 'unsaved changes' indicator.
Next, I need to confirm the deletion. I will click the 'Delete' button, then look for a confirmation prompt.
[MILESTONE: Pre-deletionChecksComplete]
The 'Delete' button is clearly visible.
</think>

Agent's Action (a_t):
<answer>
{"type": "click", "target": "[coordinates_of_delete_button]", "text": ""}
</answer>

### Applicable Principles for Evaluation (Subset of P)
Explicit Domain Principles (Human-defined):
1. EVP_SafetyGuard: "A deletion action must be confirmed via a prompt if one appears. If no prompt appears for a minor deletion, it's acceptable."
2. EVP_UIState: "Action should lead to an expected UI state change (e.g., confirmation dialog appears, or item is removed)."

Implicit Learned Principles (LLM-derived):
3. LDP_Coherence: "Reasoning should logically lead to the action and cite relevant UI cues."
4. LDP_Completeness: "Ensure all pre-conditions mentioned in the task or implied by safety are checked in reasoning before critical actions." (Relates to Goal-aligned planning)
5. LDP_Efficiency: "Avoid redundant steps if a direct action is possible."

### Reward Model Task:
Based on the agent's reasoning (c_t) and action (a_t) in the given screen state (s_t), evaluate compliance with the provided principles.
-  For LDPs: Provide a numerical score (e.g., 1-10) and a brief textual critique.
-  For EVPs: Determine if the action itself is verifiable against these rules (e.g., True/False, or a specific reward value).
\end{lstlisting}

\begin{table*}[htbp]
\centering
\caption{Orcust: Hyperparameter Configuration}
\label{tab:hyperparameters}
\
\resizebox{\textwidth}{!}{%
\begin{tabular}{
  >{\raggedright\arraybackslash}p{0.23\textwidth} 
  >{\raggedright\arraybackslash}p{0.23\textwidth} 
  >{\raggedright\arraybackslash}p{0.27\textwidth} 
  >{\raggedright\arraybackslash}p{0.23\textwidth} 
}
\toprule
\multicolumn{2}{c}{\textbf{Initial SFT Phase}} & \multicolumn{2}{c}{\textbf{RFT Phase}} \\
\cmidrule(r){1-2} \cmidrule(l){3-4}
Hyperparameter & Value & Hyperparameter & Value \\
\midrule
SFT Dataset Size             & \textasciitilde1K-5K trajectories & GRM (`r\_critique`) & Qwen2.5-72B \\
Learning Rate                & 2e-5 to 5e-6         & Trajectory Generation       & 15K (continuous)     \\
Batch Size                   & 16 - 32              & Learning Rate               & 1e-6 to 5e-7         \\
Number of Epochs             & 1 - 3                & KL Divergence Constraint    & 0.01 - 0.05          \\
Max Prompt Length            & 1024 - 2048          & Batch Size                  & 128 - 256        \\
Optimizer                    & AdamW                & Milestone Token Reward      & -0.5 / +0.5          \\
\bottomrule
\end{tabular}
}
\end{table*}

\begin{table*}[t]
\centering
\caption{Illustrative principle–reward channels employed by PCRM.  ``EVP’’ denotes \emph{Environment-Verifiable Principles}; ``LDP’’ denotes \emph{LLM-Derived Principles}.  Typical ranges refer to per–step reward contributions before weighting.}
\label{tab:principle_rewards}
\begin{tabular}{@{}p{2.2cm}p{3.0cm}p{7.9cm}p{2.2cm}@{}}
\toprule
\textbf{Category} & \textbf{Reward Type} & \textbf{Formal Check \& Rationale} & \textbf{Typical Range} \\ \midrule
\multirow{5}{*}{\textbf{EVP}} 
& Action–type correctness & $r=1$ if $\hat{a}_t.\text{type} = a_t^{\star}$, else $0$; enforces the correct primitive operation. & $\{0,1\}$ \\
& Target–bounding compliance & $r = 1-\delta$, where $\delta$ is the normalised Manhattan distance between predicted and gold coordinates (clipped at $0$); rewards spatial precision. & $[0,1]$ \\
& UI–state transition success & Binary flag from the VM harness when post–action DOM/screenshot matches an expected template (e.g.\ \emph{Settings} page visible). & $\{0,1\}$ \\
& Output-format validity & $r=-1$ if the agent’s output violates JSON/XML/span tags; prevents malformed actions. & $\{-1,0\}$ \\
& Safety guard & Immediate $r=-1$ if a destructive command is issued without prior confirmation; hard-coded safety constraint. & $\{-1,0\}$ \\ \midrule
\multirow{5}{*}{\textbf{LDP}} 
& Coherent chain-of-thought & Generative reward model $f_{\text{GRM}}$ scores whether $c_t$ sets sub-goals, cites UI cues, avoids hallucinations. & $0\!-\!10$ \\
& Goal-aligned planning & Positive score when $c_t$ explicitly references the top-level instruction and maintains it through the episode. & $0\!-\!5$ \\
& Efficiency heuristic & $-0.1$ for each redundant navigation; $+0.5$ bonus for finishing in $\le\! \tau^\star$ steps; encourages brevity. & $\approx -1\ldots+1$ \\
& User-experience etiquette & $+1$ if a risky action is preceded by confirmation in $c_t$; $-1$ if blocking pop-ups remain uncleared. & $\{-1,0,+1\}$ \\
& Cross-widget consistency & $+1$ when visual style (e.g.\ title-case) is preserved across form fields, judged by LLM critique. & $\{0,1\}$ \\ \bottomrule
\end{tabular}

\vspace{3pt}
\footnotesize\emph{Note.} Scaling factors $(w_i,v_j)$ are annealed so that EVP feedback dominates early training while LDP feedback becomes prominent after roughly $20$k steps.
\end{table*}

\subsection{Illustrative Principle-Reward Channels}
Table~\ref{tab:principle_rewards} below details some of the illustrative principle-reward channels employed by PCRM. It outlines the category of principle (EVP or LDP), the specific aspect of agent behavior being rewarded (Reward Type), the formal check or rationale behind the reward, and the typical range of the reward contribution per step (before weighting). This dual-source reward mechanism guides the agent to produce actions that are not only functionally correct but also coherent, efficient, and aligned with human preferences and task constraints.

The Orcust framework's PCRM is central to its ability to learn robust GUI interaction policies. PCRM provides stepwise and interpretable rewards by evaluating the agent's actions and reasoning against a set of predefined principles. These principles are categorized into \textbf{EVP}, which are deterministic checks against the GUI environment, and \textbf{LDP}, which leverage an LLM to critique the agent's reasoning and adherence to more nuanced, high-level strategies.

\subsection{Dataset Distribution}
Table~\ref{tab:ovtc_task_distribution_revised_wrap} presents statistical information on 15,000 high-quality multi-step GUI interaction trajectories autonomously generated via OVTC, covering various task types across different platforms. The "average steps per trajectory" refers to the number of self-labeled sub-goals or critical interaction points within each trajectory. Moreover, these trajectories consist of sequences of fundamental atomic GUI actions, with specific types detailed in Table~\ref{tab:ovtc_atomic_actions}.

\begin{table*}[h!]
    \centering
    \caption{Illustrative Task Type Distribution and Complexity in 15K OVTC Trajectories.}
    \label{tab:ovtc_task_distribution_revised_wrap}
    \begin{tabular}{@{} >{\raggedright\arraybackslash}p{2.8cm} p{0.45\textwidth} r c@{}}
        \toprule
        \textbf{Platform} & \textbf{Task Category / Example} & \textbf{Count} & \textbf{Trajectory Avg Steps} \\
        \midrule
        \textbf{Desktop Applications} & Document Editing \& Export (e.g., LibreOffice, Text Editors - save, export to PDF, change font) & 2,500 & 5 -- 8 \\
                                      & File System Operations (e.g., create folder, rename file, copy/paste) & 1,500 & 3 -- 5 \\
                                      & Email Client Usage (e.g., compose email, send, check inbox, delete email) & 1,000 & 6 -- 10 \\
                                      & Application Settings Configuration (e.g., changing preferences in a software) & 500   & 4 -- 7 \\
        \midrule
        \textbf{Web Applications}     & Online Form Filling (e.g., registration, login, contact forms) & 2,000 & 5 -- 9 \\
                                      & E-commerce Tasks (e.g., search product, add to cart, initiate checkout) & 1,500 & 6 -- 12 \\
                                      & Web Navigation \& Information Search (e.g., navigating menus, finding specific info) & 2,000 & 4 -- 8 \\
                                      & Social Media Interactions (e.g., create post, browse feed) & 500   & 4 -- 7 \\
        \midrule
        \textbf{Mobile Applications} & System Settings Adjustment (e.g., toggle Wi-Fi, change display brightness) & 1,000 & 2 -- 4 \\
                                               & Contact List Management (e.g., add contact, edit contact) & 500   & 4 -- 6 \\
                                               & Messaging App Usage (e.g., send message, read message) & 1,000 & 3 -- 6 \\
                                               & General App Navigation \& Task Completion (e.g., launching app, using core features) & 1,000 & 5 -- 10 \\
        \midrule
        \textbf{Total}                &                                                  & \textbf{15,000} &        \\
        \bottomrule
    \end{tabular}
    \par\medskip 
\end{table*}

\begin{table*}[h]
    \centering
    \caption{Illustrative Fundamental Atomic Actions in OVTC Trajectories.}
    \label{tab:ovtc_atomic_actions}
    \begin{tabular}{@{}l p{0.7\textwidth} c@{}} 
        \toprule
        \textbf{Atomic Action Type} & \textbf{Description} \\
        \midrule
        \texttt{click}                & Clicking on a UI element (button, link, menu item, coordinates).        \\
        \texttt{type} / \texttt{text\_entry}  & Entering text into a text field, search bar, or form input.      \\
        \texttt{scroll}               & Scrolling the view (vertically or horizontally) to reveal more content.  \\
        \texttt{key\_press}            & Simulating a keyboard key press (e.g., Enter, Esc, Tab, specific characters not part of a larger text entry).  \\
        \texttt{swipe}                & Swiping gestures, primarily for mobile interfaces (e.g., swipe left/right/up/down).  \\
        \texttt{drag}                 & Clicking and holding an element, then moving it to another location (e.g., drag-and-drop).  \\
        \texttt{select}               & Choosing an option from a dropdown menu or list box.               \\
        \texttt{long\_press}           & Pressing and holding an element, typically to open a context menu (mobile).   \\
        \bottomrule
    \end{tabular}
    \par\medskip
\end{table*}

\end{document}